\documentclass{article} 
\usepackage{nips14submit_e,times}
\usepackage{hyperref}
\usepackage{url}
\usepackage{cite}
\usepackage{graphicx}
\usepackage{amsmath,amssymb}
\DeclareMathOperator{\E}{\mathbb{E}}
\usepackage{subfig}
\usepackage{float}
\usepackage{epstopdf} 
\usepackage{multicol}
\usepackage{multirow}

\begin{document}

\title{Shared Loss between Generators of GANs}

\author{
Xin Wang \\
Independent Researcher\\
\texttt{xin.wang.9871@gmail.com}
}

%


\nipsfinalcopy 


\maketitle

\begin{abstract}
Generative adversarial networks are generative models that are capable of replicating the implicit probability distribution of the input data with high accuracy. Traditionally, GANs consist of a \textit{Generator} and a \textit{Discriminator} which interact with each other to produce highly realistic artificial data. Traditional GANs fall prey to the mode collapse problem, which means that they are unable to generate the different variations of data present in the  input dataset. Recently, multiple generators have been used to produce more realistic output by mitigating the mode collapse problem. We use this multiple generator framework. The novelty in this paper lies in making the generators compete against each other while interacting with the discriminator simultaneously. We show that this causes a dramatic reduction in the training time for GANs without affecting its performance.
\end{abstract}

\section{Introduction}

Generative adversarial networks (GANs)~\cite{Goodfellow14} are highly successful in replicating the implicit probability distribution of any dataset. It has been widely used in a variety of applications involving generation of realistic images ~\cite{DBLP:conf/iccv/ZhangXL17}, network analysis~\cite{saha2022machine5g, saha2022system, saha2016tv} and generation of privacy preserving synthetic datasets~\cite{dpGAN, saha2021sharks, brown2021gravitas}. The primary reason for its widespread adoption is that it produces much more realistic data than its predecessors like variational autoencoders~\cite{kingma2013auto}. 
\par
Unlike the traditional maximum likelihood estimation based generative methods, GANs are generative functions which rely on game theory. When provided with a sample, the discriminator network \textbf{\textit{D(x)}} tries to determine whether it is drawn from the generator distribution $p_{model}(x)$ or the real data distribution $p_{data}(x)$. The generator network \textbf{\textit{G(z;$\theta^{(G)}$)}} produces artificial data and tries to deceive \textbf{\textit{D(x)}} into accepting it as a real data sample. The confidence with which \textbf{\textit{D(x)}} classifies it as a real data sample is fed back to the generator as its training signal. This process continues until the game reaches a Nash equilibrium, that is $D(x)$ cannot correctly distinguish between $p_{model}(x)$ and $p_{data}(x)$.
\par
The GANs are trained using gradient descent methods which do not provide any theoretical guarantee about its convergence~\cite{DistinguishabilityCriteria}. Methods have been proposed to train GANs more efficiently~\cite{salimans2016improved} but there is still no guarantee that the training will converge $p_{model}(x)$ to $p_{data}(x)$ eventually. One of the major drawbacks of GANs is the problem of ‘mode collapse’ ~\cite{arjovsky2017towards}~\cite{che2016mode}~\cite{chen2016infogan}~\cite{metz2016unrolled}~\cite{salimans2016improved}. Theoretically, the generator is expected to learn the true data distribution at convergence. However, in practice, the generator cannot learn distributions with multiple modes correctly. For data distributions having multiple modes at $x= M_1,M_2,...,M_k$, it is observed that $p_{model}(x)$ has a single mode at $x = M_{i_1}$, such that $i_{1} \epsilon \{1,2,...,k\}$.

This problem can be addressed by improving the training methodologies of GANs~\cite{arjovsky2017towards}~\cite{metz2016unrolled}~\cite{salimans2016improved}. Another method of addressing this problem is to enforce the GAN to capture the diverse modes through various architectural modifications~\cite{DBLP:journals/corr/DurugkarGM16}~\cite{chen2016infogan}~\cite{che2016mode}. 

In this work, we propose a new architecture which trains much faster and converges much better than the existing methods. This architecture is motivated by the target to achieve better training performance and convergence in the new architectures for tackling mode-collapse.

This paper is further organized as follows: Section 2 provides a background about GANs and their working dynamics. In Section 3, we describe our motivations for pursuing this idea and its merits. Section 4 presents the methodology and detailed implementation of our idea. Section 5 includes the experimental results, followed by discussions on the merits and possible drawbacks of this scheme in Section 6. We conclude this paper in Section 7 and also suggest some directions of future work.

\section{Background}
\begin{figure}[h]
\centering
\includegraphics[width=0.8\linewidth]{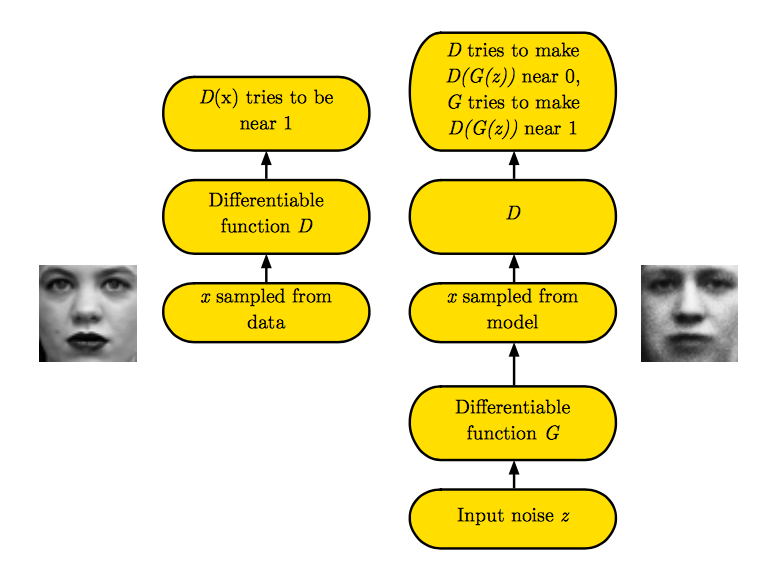}
\caption{Working of a GAN~\cite{GANtutorial}}
\label{fig:GAN_mech}
\end{figure}

Given a generator $G$ and a discriminator $D$, the training process of a GAN can be formulated as the following min-max game:
$$\min_{G} \max_{D} \E_{x \leftarrow{P_{data}}}[log D(x)] + \E_{z \leftarrow{P_{z}}}[1-log D(G(z)] $$
where $x$ is drawn from data distribution $P_{data}$ and $z$ is drawn from a prior distribution $P_z$. The mapping $G(z)$ induces a generator distribution $P_{model}$ in data space. GAN alternatively optimizes $D$ and $G$ using stochastic gradient-based learning. $G$ is therefore incentivized to map every $z$ to a single $x$ that is most likely to be classified as true data, leading to the mode collapsing problem. Another commonly asserted cause of generating less diverse samples in GAN is that, at the optimal point of $D$, minimizing $G$ is equivalent to minimizing the Jensen-Shannon divergence (JSD) between the data and model distributions, which has been empirically proven to prefer to generate samples around only a few modes whilst ignoring other modes ~\cite{DBLP:journals/corr/Huszar15}~\cite{DBLP:journals/corr/TheisOB15}. The working methodology of a GAN is depicted in Fig.\ref{fig:GAN_mech}.

\section{Motivation}
The challenge of mode collapse faced while training GANs has been tackled by using multiple generators. Various multi-generator GAN architectures ~\cite{DBLP:journals/corr/GhoshKNTD17}~\cite{DBLP:journals/corr/GhoshKN16}~\cite{LiuT16} demonstrate much better coverage of modes of the data distribution using multiple generators. It has also been theoretically proven that multiple generators are indeed successful in tackling the  mode-collapse problem~\cite{DBLP:journals/corr/abs-1708-02556}. When there are multiple generators, the discriminator not only has to guess if the sample originates from the real data distribution or the generator, but also has to determine from which generator the sample has originated. So, the discriminator forces the generators to produce different outputs from the real data distribution, thus ensuring diversity and evading mode collapse. Multiple generators are also better at producing better artificial data than GANs with a single generator~\cite{OdenaOS17}. 
\par
Other multi-agent GAN architectures also include the GMAN~\cite{DBLP:journals/corr/DurugkarGM16} architecture which involves multiple discriminators. It has been shown that using multiple discriminators can effectively train GANs in a fraction of the time required originally. However, GANs with multiple discriminators cannot address the mode-collapse problem.
\par
In this work, we propose a new class of GANs which consists of multiple generators. These generators not only learn from the training signal provided by the discriminator but also by competing against each other. We call these class of GANs Racing-GAN and show that Racing-GANs train much faster than existing GAN frameworks while tackling mode collapse at the same time. Previously, the multiple generators collaborated with each other by the means of shared weights~\cite{DBLP:journals/corr/DurugkarGM16} or by having some auxiliary information. Parallelization of GANs has also been proposed in which multiple generators are connected to multiple discriminators without any interconnection between the generators~\cite{DBLP:journals/corr/ImMKT16}. The novelty of our work lies in initiating a competition of the generators among themselves to train at a much faster rate. 

\section{Methodology}
\label{headings}

\begin{figure}[h]
\centering
\includegraphics[width=0.50\linewidth]{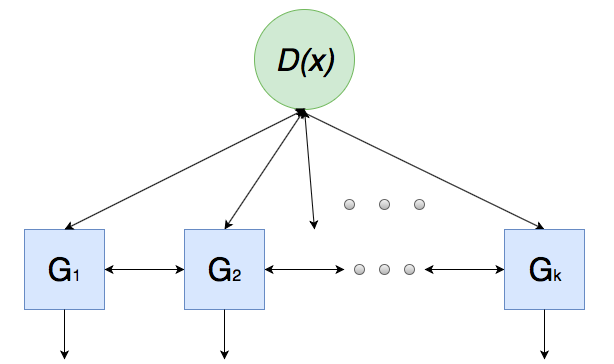}
\caption{Racing-GAN model with $k$ generators and a discriminator}
\label{fig:Flow_Chart}
\end{figure}

Racing-GAN involves a competition of multiple generators with each other. This involves an intricate interconnection of the generators among themselves. In a model with $k$ generators, $2^{k}$ such interconnections are possible. However, in this introductory paper we explore the simplest architecture possible. We connect every generator to a single discriminator. Every generator generates its own output as shown in Fig.~\ref{fig:Flow_Chart}.
In our experiments, we consider the output of any one of the generators as the final output. The choice of this generator can be random, as we show that all the generators perform equally well at the end of training. All these outputs can also be combined to produce a single output as in ~\cite{DBLP:journals/corr/abs-1708-02556}. This may include a voter which receives the outputs of all the generators and outputs the one it considers to be the best.

\subsection{Detailed Implementation}

Previously, multiple methods have been employed to make the outputs dependent on multiple generators. Some schemes utilize weight sharing ~\cite{LiuT16} while others use auxiliary information ~\cite{OdenaOS17}. In Racing-GAN, we modify the loss functions of the generators to make them compete with each other.

\par
In a model with one Discriminator and $k$ generators, the objective function of the Discriminator is:

$$\max_{D} \E_{x \leftarrow{P_{data}}}[log D(x)] + \E_{z \leftarrow{P_{z}}}[\sum_{n=1}^{k}\{1-log D(G_{n}(z)\}] $$

\par
This loss function forces the Discriminator to not only guess if the input sample belongs to the real data or not, but also to determine from which generator an artificial data sample has originated. This causes the discriminator to influence the generators to produce non-similar images, hence tackling the mode-collapse problem. The analysis of this loss function is explained in greater depth in Section 4.2.

The objective function of a generator $G_i$ in a model in which all the generators compete with each other, i.e., a fully connected model is

$$\min_{G_{i}} \E_{z \leftarrow{P_{z}}}[1-log(D(G_{i}(z) + \sum_{j \epsilon\{1,..,k\}, j \neq i}\{ max(0,D(G_{i}(z)) - D(G_{j}(z)))\}]$$

\par
In this paper, we consider a model in which a Generator $G_i$ is connected to only one Generator $G_j$. The objective function of Generator $G_i$ coupled with a single Generator $G_j \forall i,j \epsilon \{1,2,...,k\}, i \neq j $:

$$\min_{G_{i}} \E_{z \leftarrow{P_{z}}}[1-log(D(G_{i}(z) + max(0,D(G_{i}(z)) - D(G_{j}(z)))]$$

\subsection{Analysis}

The only condition for a GAN to be functional is that the loss functions should be differentiable. However, we introduce the maximum function in the loss function of the Generator but it is observed that the model works better than the original. This phenomenon can be viewed as the introduction of noise in the loss function of the GAN. Introduction of noise is a common practice in deep learning ~\cite{neelakantan2015adding} to improve learning and generalization. The objective function of the Racing-GAN can be viewed as variant of addition of noise in certain iterations. Addition of noise improves learning with gradient descent and GANs are trained using gradient descent. This intuitively explains the reason why Racing-GANs have a better training performance than the other variants of GANs. Since addition of noise during gradient descent also improves generalization, the loss functions of Racing-GANs face much lower oscillations after convergence than their counterparts.

All the generators start with random samples from the same probability distribution. It is expected that their performances will be nearly equal. But whenever generator $G_i$ performs better than generator $G_j$, $G_j$ is penalized by the addition of a positive term in its loss function. This causes $G_j$ to update its weights by a greater value, hence causing it to learn faster.

An interesting observation suggests that, due to the competitive nature of the generators in Racing-GAN, the loss functions of all the generators are closely tied to each other and tend to follow the same curve after the first few iterations. This is beneficial since it causes minimal oscillations in the loss functions after convergence.

\section{Experiments}
We present the experimental results in two parts. Firstly, we demonstrate that the Racing-GAN indeed produces good artificial data which are comparable to other GANs. Then we show that these data are produced after much less training as compared to other variants of GANs.

\subsection{Experimental Setup}
We have used a simple GAN model for our initial experiments. The generator objective is to generate a two-dimensional quadratic curve which lies between two predetermined quadratic curves. For simplicity, we use Racing-GANs with only two generators, namely $G_1$ and $G_2$.
\par
For our experiments, we use four frameworks for comparing and highlighting the efficiency of the induced competition between the generators. These frameworks are:
\begin{itemize}
    \item \textbf{GAN 1}: An ordinary GAN with only one Generator $G$
    \item \textbf{GAN 2}: A GAN with two uncoupled independent Generators $G_1$ and $G_2$
    \item \textbf{GAN 3}: A GAN with one Generator $G_2$ competing with $G_1$ whereas $G_1$ is independent of $G_2$
    \item \textbf{GAN 4 (\textit{Racing GAN})}: Both the Generators $G_1$ and $G_2$ compete with each other.
\end{itemize}

\subsection{Results}
\begin{figure}[h]
\centering
\includegraphics[width=0.9\linewidth]{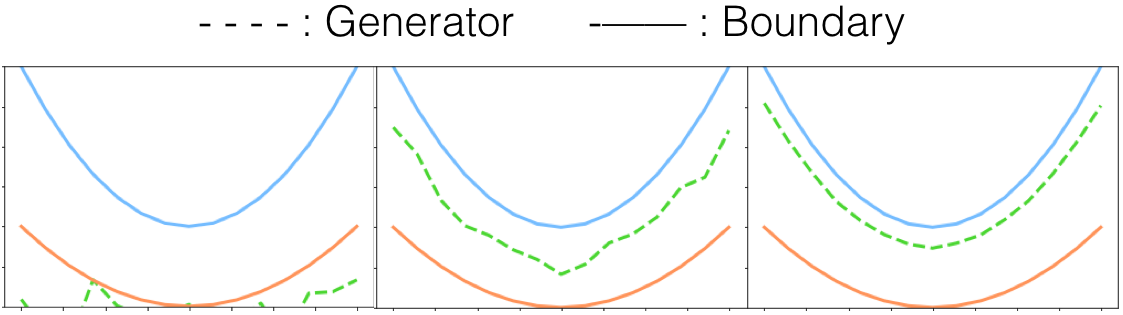}
\caption{Generator outputs at iterations  1, 2500, 8000 for an ordinary GAN with single generator }
\label{fig:Results1}
\end{figure}

\begin{figure}[h]
\centering
\includegraphics[width=0.9\linewidth]{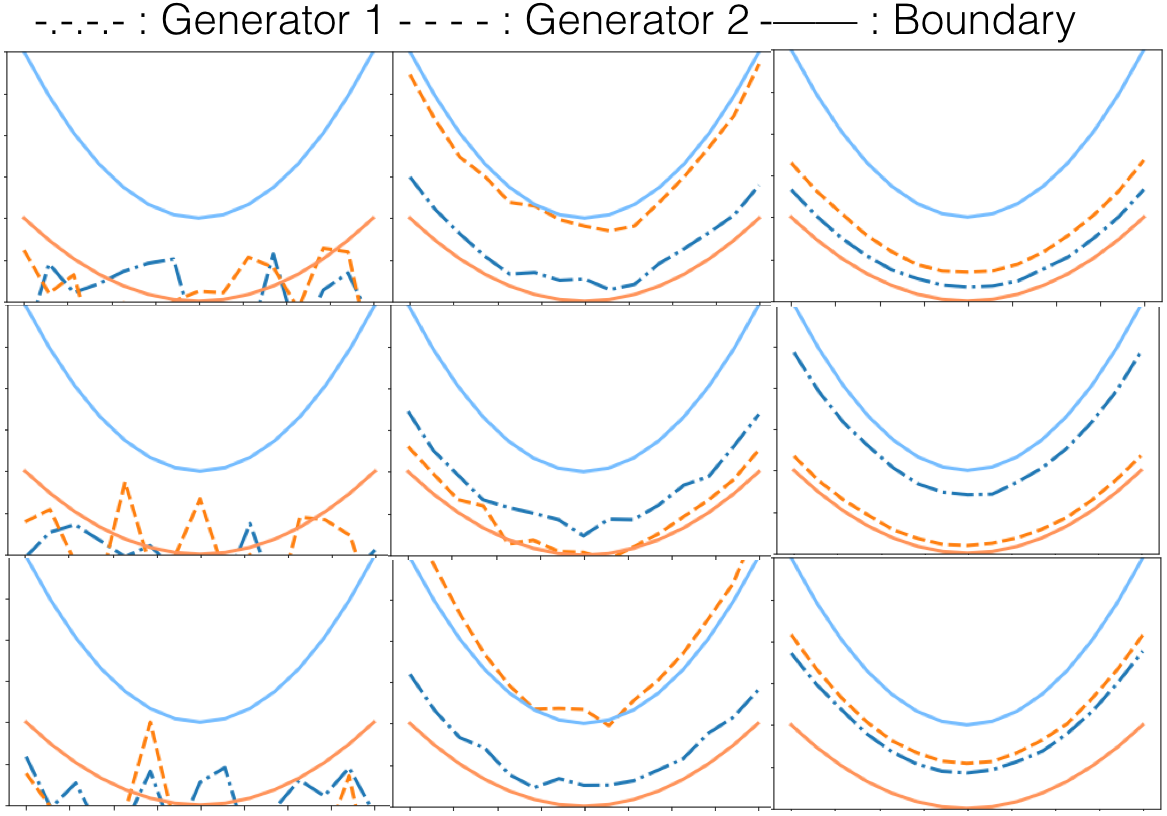}
\caption{Multiple generator outputs st iterations 1, 2500, 8000 for (from top) GAN 2, GAN 3, GAN 4}
\label{fig:Results}
\end{figure}

In this section, we present the outputs of the four variants of the GANs mentioned above after the GANs have reached their respective equilibria. They are successful in generating quadratic curves between the specified boundaries. This can be observed in the output of the iteration 8000 of the GANs in Fig.\ref{fig:Results1} and Fig.\ref{fig:Results}. From these figures, we get a rough idea that multi-generator frameworks are better than the ordinary GAN framework. This aspect is demonstrated by analyzing the convergence rate of the loss functions of these different frameworks in a later section. There we show that GAN 4
is the best among the variants considered over here.

\begin{table}[htbp]
\label{table1}
\begin{center}
\caption{Iterations required by Discriminator $D$ and Generators $G1$ and $G2$ to converge}
\begin{tabular}{llll}
\multicolumn{1}{c}{\bf GAN type}  &\multicolumn{1}{c}{\bf D}
&\multicolumn{1}{c}{\bf G}
&\multicolumn{1}{c}{\bf G1}
\\ \hline \\
GAN 1          &9348     &8107      &--\\
GAN 2          &8547    &4484       &5872\\
GAN 3          &8969    &5364       &6910\\
GAN 4          &7728    &6049       &4944\\
\end{tabular}
\end{center}
\end{table}

\begin{table}[htbp]
\label{table-2}
\begin{center}
\caption{Improvement in training efficiency over the ordinary GAN (GAN 1)  }
\begin{tabular}{lll}
\multicolumn{1}{c}{\bf GAN type}  &\multicolumn{1}{c}{\bf D}
&\multicolumn{1}{c}{\bf G}
\\ \hline \\
GAN 2          &8.57\%    &44.69\%\\
GAN 3          &4.05\%    &33.83\%\\
GAN 4          &17.33\%    &39.02\%\\
\end{tabular}
\end{center}
\end{table}

\subsection{Loss functions}
In this section, we plot the loss functions of the Discriminator and the Generators for each of the four variants of the GANs mentioned above. From Tables 1 and 2, we observe that the loss functions for the Racing-GAN converge much faster than the other variants. This shows that training of a Racing-GAN occurs faster than the other variants. (We consider that equilibrium is attained when the value of the loss function oscillates within $\pm1\%$ of its correct value.)

\begin{figure}[h]
\centering
\subfloat[GAN 1\label{fig:1a}]{\includegraphics[width=0.5\textwidth]{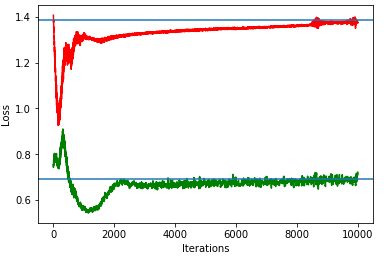}}\hfill
\subfloat[GAN 2\label{fig:1b}] {\includegraphics[width=0.5\textwidth]{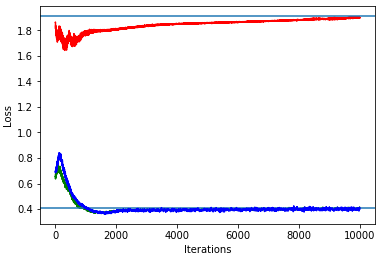}}\hfill
\subfloat[GAN 3\label{fig:1c}]{\includegraphics[width=0.5\textwidth]{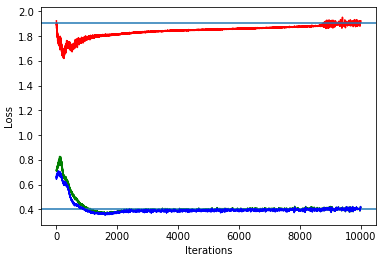}}
\subfloat[GAN 4\label{fig:1c}]{\includegraphics[width=0.5\textwidth]{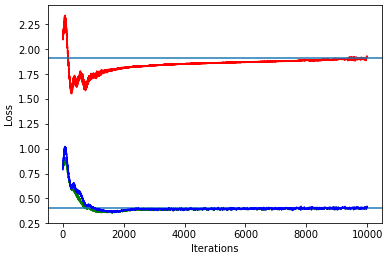}}
\caption{Loss functions of Discriminator (\textit{red}), Generator $G_1$ (\textit{green}) and Generator $G_2$ (\textit{blue})} \label{fig:1}
\end{figure}

While using the GAN to generate synthetic data, the output from any one of the generators is taken. If a multi-generator GAN is used, the generator producing the best output will be utilized. Hence, to compare fairly with other GAN variants we have used the generator which performs the best in case of multiple generators.

\subsection{Observations}
The observations stated here have been observed over several experimental iterations over the aforementioned GAN variants.
\begin{itemize}
\item From Table 2, it can be observed that a GAN with one competing generator and another independent generator (GAN 3) performs worse than a GAN with two independent generators (GAN 2).

\item In Figures 4 (a),(b),(c), it is observed that the loss functions of $G_1$ and $G_2$ closely imitate each other. This is due to the competing nature of the generators. Both of them want to be as good as the other, leading to a close overlap of their loss functions.
 
\item The generator of GAN 2 performs slightly better than that of GAN 4 but the discriminator of GAN 4 converges at least twice as fast as that of GAN 2. This has been observed in every experimental trial conducted. Hence, we can conclude that GAN 4 which is a Racing-GAN with two competing generators perform significantly better than the other variants.
\end{itemize}

\section{Discussions}
The Racing-GAN is an interesting architecture of generative adversarial networks which train much faster than its counterparts. It is expected to solve the mode collapse problem.  We expect to see an increased interest in the research community where various generators are combined in different fashions to get better results. The Racing-GAN is an approximately differentiable model, which is inspired from nature, seems to work better than the other variants. However, one of the basic requirements for GANs to operate is that the loss function has to be differentiable. This is a contradiction which is unsolved as of now. An interpretation of the success of Racing-Gan in spite of being non-differentiable is that the extra term being added to the generator loss function is being treated as noise. It has been shown that addition of noise improves generalization and avoids overfitting in deep neural networks\cite{neelakantan2015adding}. 

\section{Conclusion and Future Work}
We present a new model of GAN called a Racing-GAN which trains much faster and effectively than the existing variants of GANs. This model has multiple generators which compete with each other besides learning from the training signal provided by the discriminator. The work presented in this paper is of practical and experimental nature. We hope to develop a deeper theoretical understanding of its dynamics in the near future. Other future work might also include investigating various couplings between the generators and modifying the loss functions for better results.

\bibliography{output}
\bibliographystyle{IEEEtran}

\end{document}